\documentclass[11pt]{article}

\usepackage[final]{acl}

\usepackage{times}
\usepackage{latexsym}

\usepackage[T1]{fontenc}
\usepackage[utf8]{inputenc}

\usepackage{microtype}
\usepackage{inconsolata}

\usepackage{graphicx}

\usepackage{booktabs}
\usepackage{tabularx}
\usepackage{multirow}
\usepackage{colortbl}
\usepackage{subcaption}

\usepackage{amsmath}
\usepackage{amssymb}
\usepackage{amsfonts}
\usepackage{nicefrac}

\usepackage{algorithm}
\usepackage{algpseudocode}

\usepackage{enumitem}
\usepackage{xcolor}
\usepackage{tcolorbox}
\tcbuselibrary{breakable}

\usepackage{url}

\title{HAGE: Harnessing Agentic Memory via RL-Driven Weighted Graph Evolution}

\author{
Dongming Jiang$^{\alpha}$, Yi Li$^{\alpha}$, Guanpeng Li$^{\beta}$, Qiannan Li$^{\gamma}$, Bingzhe Li$^{\alpha}$\thanks{Corresponding author} \\
$^{\alpha}$Department of Computer Science, The University of Texas at Dallas \\
$^{\beta}$Department of Electrical and Computer Engineering, University of Florida \\
$^{\gamma}$University of California, Davis \\
\texttt{\{dongming.jiang, yi.li3, bingzhe.li\}@utdallas.edu} \\
\texttt{liguanpeng@ufl.edu} \\
\texttt{qnli@ucdavis.edu}
}

\begin{document}
\maketitle

\begin{abstract}
Memory retrieval in agentic large language model (LLM) systems is often treated as a static lookup problem, relying on flat vector search or fixed binary relational graphs. However, fixed graph structures cannot capture the varying strength, confidence, and query-dependent relevance of relationships between events. In this paper, we propose HAGE, a weighted multi-relational memory framework that reconceptualizes retrieval as sequential, query-conditioned traversal over a unified relational memory graph. Memory is organized as relation-specific graph views over shared memory nodes, where each edge is associated with a trainable relation feature vector encoding multiple relational signals. Given a query, an LLM-based classifier identifies the relational intent, and a routing network dynamically modulates the corresponding dimensions of the edge embedding. Traversal scores are computed via a learned combination of semantic similarity and these query-conditioned edge representations. This allows memory traversal to prioritize high-utility relational paths while softly suppressing noisy or weakly relevant connections. Beyond adaptive traversal, HAGE further introduces a reinforcement learning-based training framework that jointly optimizes routing behavior and edge representations using downstream tasks. Finally, empirical results demonstrate improved long-horizon reasoning accuracy and a favorable accuracy-efficiency trade-off compared to state-of-the-art agentic memory systems. Our code is available at \url{https://github.com/FredJiang0324/HAGE_MVPReview}.
\end{abstract}

\section{Introduction}
\label{sec:intro}
Large Language Models (LLMs) have rapidly become the foundation of modern AI agents~\citep{brown2020language,achiam2023gpt,wei2022chain,yao2022react,shinn2023reflexion,park2023generative}, enabling strong performance in reasoning, planning, tool use, and multi-turn interaction~\citep{Brown2020,achiam2023gpt,Wei2022}. However, effective agency requires more than solving isolated prompts. A long-horizon agent must accumulate experience, retain user- and task-specific information, and selectively reuse past evidence across sessions. This requirement exposes a fundamental limitation of context-only interaction: even when long-context models are available, relevant information can be diluted, misplaced, or forgotten as interactions grow, leading to unstable recall and degraded long-term reasoning~\citep{liu2024lost,Beltagy2020,maharana2024evaluating,wu2024longmemeval}.

Retrieval-Augmented Generation (RAG) and memory-augmented generation systems address this issue by moving part of the agent's knowledge outside the model parameters and into an explicit, queryable memory store~\citep{lewis2020retrieval,borgeaud2022improving,packer2024memgptllmsoperatingsystems,zhong2024memorybank}. Such external memories allow agents to preserve information beyond the current context window, support multi-session continuity, and adapt responses based on accumulated experience. Recent agent-memory systems further move beyond simple document retrieval by extracting salient memories, updating them over time, and organizing them into structured representations such as episodic records, semantic summaries, entity-centric memories, or graph-based links~\citep{xu2025mem,chhikara2025mem0}. These designs show that the structure of memory is crucial for long-term agent behavior.

Despite this progress in structuring memory, a central challenge remains underexplored: \emph{how should an agent prioritize and navigate these complex connections?}
Graph-based memory and graph-augmented retrieval have emerged as promising directions for capturing semantic, temporal, causal, and entity-centric dependencies in complex reasoning tasks~\citep{edge2024graphrag,gutierrez2024hipporag,rasmussen2025zep,anokhin2024arigraph}.
However, most existing agent-memory approaches still rely on unweighted or weakly weighted relations, where an edge primarily indicates the existence of a connection rather than its query-dependent utility.
This is a critical bottleneck. In real-world reasoning, the importance of a connection is inherently query-dependent. For example, a temporal edge might be essential for answering a sequence-based question but irrelevant for an entity-centric query. By treating outgoing connections as equally valid or using fixed graph-expansion rules, existing systems can fail to discriminate between highly relevant pathways and distracting noise, leading to degraded retrieval accuracy as memory grows.

Furthermore, even when continuous scores or edge weights are introduced, retrieval is still largely governed by fixed similarity search, manually designed scoring functions, or static heuristic traversal rules.
Recent work on adaptive RAG and graph-based retrieval suggests that retrieval decisions can be optimized through learned policies or reinforcement learning rather than predefined pipelines~\citep{guo2025routerag,yu2026graphrag}.
However, these methods mainly target external knowledge-intensive QA or text-graph hybrid retrieval, rather than persistent agentic memory where the memory graph evolves across interactions.
This gap motivates a shift toward dynamic routing for agentic memory: instead of relying on handcrafted access mechanisms, an agent should learn which relational paths to follow based on the immediate query and downstream feedback.

To address these limitations, we propose HAGE, a weighted multi-relational memory framework that reconceptualizes memory retrieval as query-conditioned traversal over a multi-relational memory graph with relation-specific views, trained with reinforcement learning-based optimization. HAGE is built on two key principles.

First, memory is structured as a family of relation-specific graphs with trainable edge embeddings. Instead of static scalar weights, each embedding encodes multiple relational dimensions. Given a query, an LLM-based classifier identifies the relational intent, and a routing network dynamically modulates these edge features. By additively combining semantic similarity with this query-conditioned structural weight, the system respects both content relevance and structural alignment. This design enables query-dependent routing, allowing the agent to efficiently traverse structurally critical but semantically distant bridge nodes.

Second, HAGE introduces a reinforcement learning-based training framework for adaptive retrieval. Instead of relying on fixed traversal heuristics, the model learns to optimize relation-aware routing behavior using downstream task feedback. In our formulation, trainable edge representations capture which relational connections are useful for different query types, while the routing component determines how retrieval proceeds conditioned on the query. This coupling allows the retrieval policy and memory representations to be optimized jointly, yielding a learned alternative to handcrafted graph traversal strategies.

Together, these contributions shift agentic memory from fixed heuristic retrieval toward learned relation-aware retrieval. Instead of relying solely on manually designed graph scoring rules, HAGE treats retrieval as an optimized, query-conditioned traversal process over a multi-relational memory graph.

Our contributions are summarized as follows:
\begin{enumerate}[leftmargin=*, itemsep=0pt, topsep=2pt]
    \item A weighted multi-relational memory architecture in which a multi-relational memory graph is augmented with learnable edge representations, enabling fine-grained, per-edge discrimination beyond static or type-level heuristic scoring.

    \item A reinforcement learning framework that formulates query-conditioned graph retrieval as a sequential decision process. It jointly optimizes routing behavior and edge representations using downstream task feedback, requiring only node-level evidence targets rather than full path-level trajectory supervision.

    \item An empirical analysis showing that joint optimization with regularization improves generalization over routing-only and edge-only variants, highlighting the importance of learned edge representations for robust graph-based memory retrieval.\footnote{The MVP implementation has been open-sourced at: \url{https://github.com/FredJiang0324/HAGE_MVPReview}.}
\end{enumerate}

\begin{figure}[t]
    \centering
    \includegraphics[width=\columnwidth]{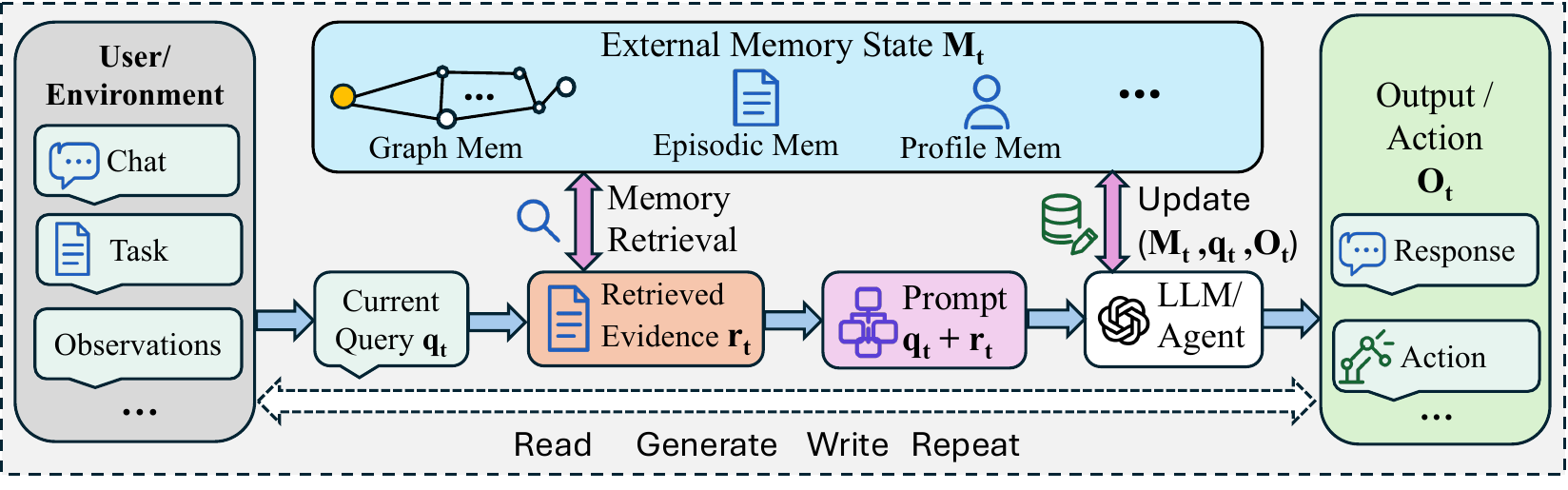}
    \caption{High-Level Architecture of Memory-Augmented Generation (MAG).}
    \label{fig:mag_overview}
\end{figure}

\section{Background}
\label{sec:related}
\subsection{From Static Retrieval to Agentic Memory}
Retrieval-Augmented Generation (RAG) improves language models by retrieving relevant information from an external datastore and conditioning generation on the retrieved context~\citep{lewis2020retrieval}. While this paradigm is effective for relatively static corpora, long-horizon agents require a more dynamic form of retrieval: they must accumulate, update, and reuse information generated through their own interactions. This motivates Memory-Augmented Generation (MAG) as shown in Figure~\ref{fig:mag_overview}, where the memory store is not only queried but also revised over time as the agent observes new events, user preferences, task outcomes, and environmental feedback~\citep{park2023generative,packer2024memgptllmsoperatingsystems,nan2025nemori,chhikara2025mem0,xu2025mem}.

Formally, at interaction step $t$, an agent maintains a mutable memory state $\mathcal{M}_t$. Given a query or observation $q_t$, the agent retrieves relevant evidence from memory, generates an output, and then updates the memory state:
\begin{equation}
    r_t = \mathrm{Retrieve}(q_t, \mathcal{M}_t),
\end{equation}
\begin{equation}
    o_t = \mathrm{LLM}(q_t, r_t),
\end{equation}
\begin{equation}
    \mathcal{M}_{t+1} = \mathrm{Update}(\mathcal{M}_t, q_t, o_t).
\end{equation}
This read--generate--write loop distinguishes agentic memory from conventional retrieval. The memory system must not only preserve useful information, but also determine how relevant evidence should be accessed.

Recent work has explored increasingly structured forms of agent memory, including episodic summaries, note-like memory units, entity-centered memory stores, and graph-based relational memories~\citep{liu2023think,xu2025mem,nan2025nemori,edge2024graphrag,rasmussen2025zep,kiciman2023causal}. Graph-based memory is particularly appealing because it can encode semantic, temporal, causal, and entity relations explicitly, allowing retrieval to exploit relational structure instead of relying only on embedding similarity. However, in many such systems, memory access still depends on fixed edge types, manually designed weighting rules, or heuristic traversal procedures. Thus, although the memory representation becomes more expressive, the access mechanism often remains static.

\subsection{Learning Memory Access as Sequential Decision Making}

HAGE focuses on this underexplored problem: how to learn the retrieval behavior of a structured memory system. We view graph-based memory access as a sequential decision process. Given a query and the current memory graph, the system must decide which neighbors to expand, which relational cues to emphasize, and which memory nodes to include in the retrieved context. This formulation is particularly natural for multi-hop, temporal, and causal queries, where the usefulness of a memory item depends not only on its individual relevance but also on the path through which it is reached.

This perspective connects graph-based memory retrieval with reinforcement learning. Rather than treating traversal as a fixed procedure, one can optimize retrieval decisions using rewards derived from downstream evidence quality. HAGE adopts this view by making both edge representations and routing behavior trainable. Edge features capture relation-aware traversal preferences, while the routing policy learns how to traverse the graph under task-level feedback. In this way, memory structure and memory access are optimized jointly rather than designed independently.

\section{HAGE Design}
In this section, we introduce HAGE, a framework that reconceptualizes memory retrieval in agentic systems as sequential, query-conditioned traversal over structured relational memory, rather than as static lookup. HAGE consists of two key components: (1) a weighted multi-relational graph memory for capturing heterogeneous and strength-sensitive relations among memory events, and (2) a reinforcement learning-based training framework for jointly optimizing relation-aware retrieval policies and edge representations. We first present the construction of the weighted graph memory and its query-conditioned traversal mechanism, followed by the learning framework used to optimize routing behavior and relational edge weights.

\begin{figure*}[t]
    \centering
    \includegraphics[width=0.8\linewidth]{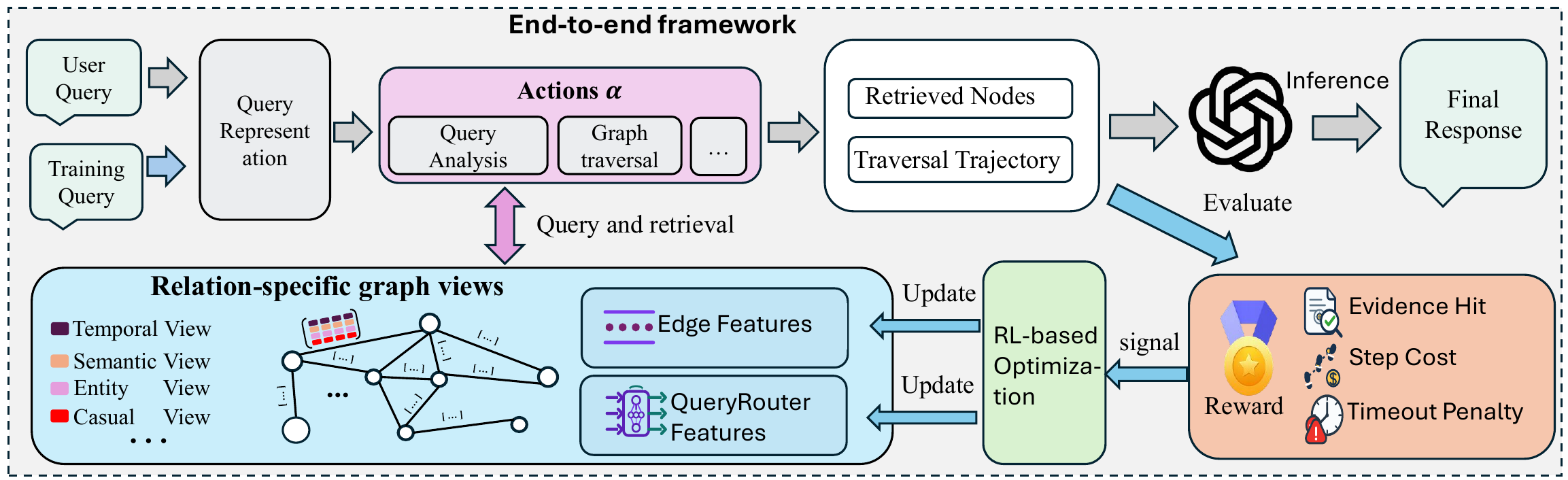}
    \caption{Architectural Overview of HAGE.}
    \label{fig:coevo_overview}
\end{figure*}

\subsection{Overview}
\label{sec:design:overview}
HAGE is built on the insight that memory retrieval in agentic systems requires more than static lookup: it often involves sequential, query-conditioned traversal over structured memory. To operationalize this perspective, HAGE integrates two tightly coupled components, as illustrated in Figure~\ref{fig:coevo_overview}.

\begin{itemize}[leftmargin=*, itemsep=0pt, topsep=2pt]
    \item A \textbf{weighted multi-relational memory graph}, where each edge carries a trainable feature vector encoding relation-aware traversal preferences. These features are initialized from a heuristic scoring phase and refined through downstream reward signals.
    \item A \textbf{reinforcement learning-based training framework} that jointly optimizes a query-conditioned routing network and the edge representations using policy-gradient updates.
\end{itemize}

Unlike prior graph-based memory systems that rely on fixed edge types and hand-designed scoring rules, HAGE makes relation weighting query-adaptive and learnable.

\subsection{Weighted Multi-Relational Memory Graph}
\label{sec:design:graph}
We represent memory as a directed multigraph $\mathcal{G}_t = (\mathcal{N}_t, \mathcal{E}_t)$. The edge set is decomposed into four relation-specific subsets that capture temporal adjacency, semantic similarity, causal dependence, and entity co-reference:
\begin{equation}
\mathcal{E}_t = \mathcal{E}_{temp} \cup \mathcal{E}_{sem} \cup \mathcal{E}_{causal} \cup \mathcal{E}_{ent}.
\end{equation}
Nodes are hierarchically organized into fine-grained \textit{Event-Nodes}. Each Event-Node $n_i$ is represented as
\begin{equation}
n_i = \langle c_i, \tau_i, \mathbf{v}_i, \mathcal{A}_i \rangle,
\end{equation}
where $c_i$ denotes the event content, $\tau_i$ is the associated timestamp, $\mathbf{v}_i \in \mathbb{R}^d$ is a dense semantic embedding, and $\mathcal{A}_i$ contains structured metadata associated with the event.

A key design choice in HAGE is that each edge $(i,j)$ is associated with a trainable relation feature vector $\mathbf{e}_{ij} \in \mathbb{R}^{R}$, where $R=4$ in this design, corresponding to temporal, semantic, causal, and entity-based relations. When an LLM-based edge-scoring cache is available, we initialize this vector as
\begin{equation}
\mathbf{e}_{ij}^{(0)} = \left[ s_{temp}, \; s_{sem}, \; s_{causal}, \; s_{ent} \right]^\top,
\end{equation}
where $s_r$ denotes the initial score assigned to relation type $r$. In the absence of cached scores, $\mathbf{e}_{ij}^{(0)}$ is initialized as a one-hot vector corresponding to the edge's primary relation type. During training, these edge features are optimized as learnable parameters and updated using downstream reward signals.

\subsection{Query-Conditioned Retrieval}
\label{sec:design:query}

Given a query $q$ and graph $\mathcal{G}_t$, HAGE performs retrieval in four stages: query analysis, anchor identification, weighted traversal, and context synthesis.

\paragraph{Query analysis and anchor identification.}
The query is mapped to structured control signals, including a relation intent $T_q$, a dense embedding $\vec{q}$, and auxiliary lexical or temporal constraints when available. To initialize traversal robustly, the system identifies anchor nodes by fusing multiple retrieval signals, including dense vector retrieval, sparse lexical matching, and temporal filtering. In practice, this stage provides reliable entry points, while the core contribution of HAGE lies in the learned traversal that follows.

\paragraph{Query-conditioned weighted traversal.}
Starting from the anchor set $\mathcal{S}_{anchor}$, the system expands the retrieved context through weighted graph traversal. For a given query $q$, let $\mathbf{v}_{T_q}$ denote the dense embedding of the relation intent $T_q$ identified by the LLM-based classifier. For each edge $(i,j)$, the static feature $\mathbf{e}_{ij}$ is augmented with runtime similarity features and the query intent:
\begin{equation}
\tilde{\mathbf{e}}_{ij} = \left[ \mathbf{e}_{ij}; \; \mathbf{v}_{T_q}; \; \cos(\vec{q}, \mathbf{v}_i); \; \cos(\vec{q}, \mathbf{v}_j) \right].
\end{equation}
The enriched feature and query embedding are passed through a lightweight MLP, denoted \textit{QueryRouter}, which produces a positive scalar structural weight:
\begin{equation}
w_{ij}(q) = \mathrm{softplus}\!\left(\mathrm{MLP}\!\left([\vec{q}; \; \tilde{\mathbf{e}}_{ij}]\right)\right).
\end{equation}
To ensure the agent can traverse structurally critical but semantically distant ``bridge'' nodes, the final transition score is defined as an additive combination of semantic relevance and the learned structural weight:
\begin{equation}
S(n_j \mid n_i, q) = \lambda \cos(\mathbf{v}_j, \vec{q}) + (1 - \lambda) w_{ij}(q),
\end{equation}
where $\lambda \in [0, 1]$ is a balancing hyperparameter. This additive form ensures that an edge can be strongly preferred if it possesses high structural importance, even if the target node has a negative semantic cosine similarity. The resulting traversal policy is
\begin{equation}
\pi(n_j \mid n_i, q) = \frac{\exp(S(n_j \mid n_i, q))}{\sum_{n_k \in \mathcal{N}(n_i)} \exp(S(n_k \mid n_i, q))},
\end{equation}
where $\mathcal{N}(n_i)$ denotes the neighbors of $n_i$. During training, actions are sampled from $\pi$ for exploration; at inference time, the system uses greedy selection or beam-style expansion over high-scoring candidates. Traversal terminates when the hop budget is exhausted or target evidence is reached.

\paragraph{Context synthesis.}
The retrieved nodes are reordered and serialized into a compact context for the downstream LLM. Depending on query type, nodes are organized temporally, causally, or by retrieval score, and are included until the context budget is exhausted.

\subsection{Reinforcement Learning-Based Joint Optimization}
\label{sec:design:rl}

HAGE optimizes relation-aware retrieval by formulating graph traversal as a Markov Decision Process (MDP) and training the routing network and edge representations jointly via policy gradient methods.

\paragraph{MDP Formulation.} Each training example defines a per-query episode:
\begin{itemize}[leftmargin=*, itemsep=0pt, topsep=2pt]
    \item \textbf{State:} The current node $n_i$, the query embedding $\vec{q}$, and a visited-node mask $\mathcal{V}_t$ to prevent cyclic loops.
    \item \textbf{Action:} Selecting a neighbor $n_j \in \mathcal{N}(n_i)$ according to the stochastic policy $\pi_\theta(n_j \mid n_i, q)$.
    \item \textbf{Transition:} The agent moves to $n_j$ and the step count increments.
    \item \textbf{Termination:} The episode ends when the agent reaches a target evidence node, encounters a dead end (no unvisited neighbors), or exhausts the hop budget $H_{max}$.
\end{itemize}
The start node is selected as the node with highest cosine similarity to the query embedding, simulating the anchor identification stage during training.

\paragraph{Reward Design.} The reward combines an evidence-hit signal with shaping penalties for traversal cost:
\begin{equation}
r_t = r_t^{hit} - \lambda_{step} r_t^{step} - \lambda_{timeout} r_t^{timeout},
\label{eq:reward}
\end{equation}
where $r_t^{hit}$ rewards retrieving target evidence nodes (identified during training by matching node content with ground-truth answers). For multi-hop queries, the agent accumulates $r_t^{hit}$ for each unique target found; traversal terminates only when all required evidence is collected, a dead end is reached, or the hop budget is exhausted. Lastly, $r_t^{step}$ and $r_t^{timeout}$ penalize excessive hops and budget exhaustion, encouraging the model to discover efficient, direct relational paths.

\paragraph{Policy Gradient with Baseline Subtraction.} We optimize the traversal policy using REINFORCE with an exponential moving average baseline for variance reduction. For a trajectory $\tau = (n_0, a_0, r_0, \ldots, n_T)$, the discounted return at step $t$ is
\begin{equation}
G_t = \sum_{k=0}^{T-t} \gamma^k r_{t+k},
\end{equation}
where $\gamma$ is the discount factor. The policy-gradient update is
\begin{equation}
\nabla_\theta \mathcal{J} = \sum_{t=0}^{T} \nabla_\theta \log \pi_\theta(a_t \mid s_t)\cdot (G_t - b),
\label{eq:reinforce}
\end{equation}
where $b$ is a running baseline updated using exponential moving averaging. The parameter set $\theta$ includes both the QueryRouter weights and the trainable edge features, allowing the two components to be optimized under the same reward signal. Gradients are clipped to improve stability.

\paragraph{Anchor Regularization.} Since the edge features are warm-started from Phase~1 scores, unconstrained optimization may cause them to drift far from their initial values. This creates a distribution mismatch at inference: unseen graphs use static Phase~1 features, while the router was trained on drifted features. To prevent this, we add an L2 anchor regularization term:
\begin{equation}
\mathcal{L}_{anchor} = \lambda_{anchor} \sum_{(i,j) \in \mathcal{E}_{train}} \left\| \mathbf{e}_{ij} - \mathbf{e}_{ij}^{(0)} \right\|_2^2,
\label{eq:anchor}
\end{equation}
where $\mathbf{e}_{ij}^{(0)}$ denotes the frozen Phase~1 initialization. The total training objective combines the policy gradient with this regularization:
\begin{equation}
\mathcal{L} = -\mathcal{J}(\theta) + \mathcal{L}_{anchor}.
\end{equation}

This formulation can be interpreted as a form of constrained policy learning, where exploration in the feature space is explicitly regularized toward a semantically meaningful initialization, enabling robust generalization to new memory graphs.

\subsubsection{Co-Evolutionary Training Dynamics}

The joint optimization creates a co-evolutionary dynamic between two parameter groups:
\begin{itemize}[leftmargin=*, itemsep=0pt, topsep=2pt]
    \item \textbf{Edge features} ($\mathbf{e}_{ij}$) adapt to encode traversal-relevant signals that the router can exploit. Features on successful trajectories are reinforced, while those on unsuccessful paths are suppressed.
    \item \textbf{QueryRouter weights} learn to map query--edge feature pairs to traversal preferences, discovering which feature patterns predict useful transitions for different query types.
\end{itemize}

To stabilize this feedback-driven co-evolution, we use asymmetric learning rates: $\eta_{router}$ for the QueryRouter and $\eta_{edge}<\eta_{router}$ for the edge features. This allows the router to adapt rapidly to query-conditioned traversal preferences, while edge features evolve more conservatively to preserve the Phase~1 semantic structure and avoid unstable feature drift.

\subsection{Implementation}
\label{sec:implementation}

HAGE is implemented in PyTorch as a modular graph-based training framework. Each memory graph is represented using node embeddings, COO-format edge indices, typed edge labels, and relation-specific edge features, enabling GPU-accelerated routing and edge optimization. We use all-MiniLM-L6-v2~\citep{reimers2019sentence} to initialize node embeddings and precompute adjacency lists for efficient traversal.

Training is performed with sample-level cross-validation. The router and edge modules are optimized with Adam~\citep{kingma2014adam}, using separate learning rates for routing and edge-feature updates. The best checkpoint is selected based on validation routing success rate. Importantly, Phase~2 training requires no LLM calls, operating only on cached graph structures and pre-computed embeddings. Detailed hyperparameters are provided in Appendix~\ref{app:implementation}.

\begin{table*}[t]
\centering
\small
\caption{LoCoMo comparison with LLM-as-a-judge score under different methods and backbone LLMs. Higher is better. Best results are shown in \textbf{bold}, and second-best results are \underline{underlined}. HAGE is our proposed method.}
\label{tab:locomo-main}
\resizebox{0.8\textwidth}{!}{
\begin{tabular}{lccccc|c}
\toprule
Method &
Multi-Hop &
Temporal &
Open-Domain &
Single-Hop &
Adversarial &
Overall \\
\midrule
\multicolumn{7}{c}{\textit{gpt-4o-mini}} \\
\midrule

Full Context &
0.468 &
0.562 &
0.486 &
0.630 &
0.205 &
0.481 \\

A-MEM &
0.495 &
0.474 &
0.385 &
0.653 &
0.616 &
0.580 \\

MemoryOS &
\underline{0.552} &
0.422 &
\underline{0.504} &
0.674 &
0.428 &
0.553 \\

Nemori &
\textbf{0.569} &
0.649 &
0.485 &
0.764 &
0.325 &
0.590 \\

MAGMA &
0.528 &
\underline{0.650} &
\textbf{0.517} &
\underline{0.776} &
\underline{0.742} &
\underline{0.700} \\

MemSkill &
0.480 &
0.453 &
0.498 &
0.614 &
0.317 &
0.501 \\

\textbf{HAGE (ours)} &
0.547 &
\textbf{0.667} &
0.497 &
\textbf{0.797} &
\textbf{0.839} &
\textbf{0.739} \\

\midrule
\multicolumn{7}{c}{\textit{Qwen2.5-3B}} \\
\midrule

Full Context &
0.229 &
0.095 &
\underline{0.335} &
0.227 &
0.244 &
0.215 \\

A-MEM &
0.258 &
0.203 &
0.219 &
0.416 &
\textbf{0.684} &
0.410 \\

MemoryOS &
0.285 &
0.212 &
0.194 &
0.341 &
0.229 &
0.280 \\

Nemori &
\textbf{0.317} &
\underline{0.450} &
\textbf{0.379} &
\underline{0.641} &
0.036 &
0.412 \\

MAGMA &
0.301 &
0.402 &
0.334 &
0.576 &
0.589 &
\underline{0.499} \\

MemSkill &
0.149 &
0.079 &
0.158 &
0.187 &
0.266 &
0.179 \\

\textbf{HAGE (ours)} &
\underline{0.315} &
\textbf{0.457} &
\underline{0.335} &
\textbf{0.657} &
\underline{0.603} &
\textbf{0.548} \\

\bottomrule
\end{tabular}}
\end{table*}

\section{Experiments}
\label{experiments}
We conduct comprehensive experiments to evaluate the proposed HAGE architecture, focusing on three aspects: (1) end-to-end reasoning accuracy on long-term memory benchmarks, (2) the effectiveness of co-evolutionary edge learning via ablation studies, and (3) system efficiency under realistic deployment conditions.

\subsection{Experimental Setup}
\label{sec:exp_setup}

\noindent\textbf{Datasets.}
We evaluate memory retrieval capability using two widely adopted benchmarks: \textbf{(1) LoCoMo}~\citep{maharana2024evaluating}: which contains ultra-long conversations (average length of 9K tokens) designed to assess long-range temporal and causal retrieval.
\textbf{(2) HotpotQA}~\citep{yang2018hotpotqa}: a multi-hop question answering benchmark requiring reasoning over multiple supporting facts. We use it to evaluate whether the memory retriever can identify and connect dispersed evidence across documents, thereby testing cross-evidence retrieval and compositional reasoning capability.

\noindent\textbf{Baselines.}
We compare HAGE with Full Context and five state-of-the-art memory architectures using the same backbone LLMs:
\begin{description}[leftmargin=!, labelwidth=1.8cm, itemsep=0pt, topsep=1pt]
    \item[Full Context.] Feeds the entire conversation history into the LLM.
    \item[A-MEM~\citep{xu2025mem}.] A self-evolving agent memory system.
    \item[Nemori~\citep{nan2025nemori}.] A graph-based memory with predict-calibrate episodic segmentation.
    \item[MemoryOS~\citep{Kang2025}.] A hierarchical semantic memory operating system.
    \item[MAGMA~\citep{jiang2026magma}.] A multi-relational memory with static edge weights and heuristic traversal.
    \item[MemSkill~\citep{zhang2026memskill}.] An RL-based skill-evolving memory method.
\end{description}

\noindent\textbf{Metrics.}
Our primary metric is the LLM-as-a-Judge score~\citep{zheng2023judging}, which evaluates semantic correctness through an instruction-tuned model (prompt details in the appendix). We additionally report token-level F1 as supplementary lexical measures.

\noindent\textbf{Evaluation Protocol.}
For the RL-trained components, including trainable edge features and the query router, we adopt a 5-fold cross-validation protocol at the conversation-sample level. This ensures that all queries from the same conversation sample are kept within the same split, preventing query-level leakage across training and evaluation. Each sample is evaluated exactly once by a model that has not observed it during training. We report the mean across folds.

\noindent\textbf{Training Configuration.}
We use the same locked training configuration across all folds and select checkpoints based only on validation reward. Detailed hyperparameters are provided in Appendix~\ref{app:implementation}.

\subsection{Overall Performance on LoCoMo}

We first evaluate HAGE on LoCoMo, a long-term conversational memory benchmark. Table~\ref{tab:locomo-main} reports the results under two backbone LLMs: gpt-4o-mini and Qwen2.5-3B. HAGE achieves the best overall performance under both backbone settings. With gpt-4o-mini, HAGE improves the overall judge score from the strongest baseline score of $0.700$ to $0.739$. With Qwen2.5-3B, HAGE improves the strongest baseline score from $0.499$ to $0.548$. These results show that HAGE provides consistent gains across both stronger and smaller backbone models.

A closer analysis shows that HAGE is particularly effective on reasoning-intensive categories. Under gpt-4o-mini, HAGE achieves the best scores on Temporal, Single-Hop, Adversarial, and Overall categories, with especially large gains on Adversarial queries. Under Qwen2.5-3B, HAGE achieves the best scores on Temporal, Single-Hop, and Overall categories. These gains suggest that learned query-adaptive traversal can help retrieve more useful evidence before answer generation, reducing the burden on the backbone LLM.

\subsection{Generalization to Non-Conversational Multi-Hop QA}

To evaluate whether HAGE generalizes beyond long-term conversational memory, we further evaluate it on HotpotQA under the distractor setting. Unlike LoCoMo, HotpotQA is a non-conversational multi-hop question answering benchmark that requires identifying and combining supporting evidence from multiple distractor passages. This setting provides a complementary testbed for evidence-intensive multi-hop reasoning.

As shown in Table~\ref{tab:hotpotqa}, HAGE achieves the best overall performance on HotpotQA under the distractor setting, obtaining the F1 score of 0.678 and the LLM score of 0.824 with GPT-4o-mini. The same trend also holds for Qwen2.5-3B, where HAGE consistently outperforms all baselines. These results indicate that HAGE's learned traversal mechanism can generalize beyond conversational memory and remain effective in non-conversational multi-hop reasoning settings.

\begin{table}[t]
\centering
\small
\setlength{\tabcolsep}{5pt}
\caption{HotpotQA comparison with F1 and LLM score under the distractor setting. Higher is better. Best results are shown in \textbf{bold}, and second-best results are underlined.}
\label{tab:hotpotqa}
\begin{tabular}{lcccc}
\toprule
\multirow{2}{*}{\textbf{Method}}
& \multicolumn{2}{c}{\textbf{GPT-4o-mini}}
& \multicolumn{2}{c}{\textbf{Qwen2.5-3B}} \\
\cmidrule(lr){2-3} \cmidrule(lr){4-5}
& \textbf{F1} & \textbf{LLM Score} & \textbf{F1} & \textbf{LLM Score} \\
\midrule
A-MEM      & 0.433 & 0.547 & 0.186 & 0.416 \\
MemoryOS   & 0.477 & 0.592 & \underline{0.350} & \underline{0.459} \\
Nemori     & 0.131 & 0.624 & 0.091 & 0.332 \\
MAGMA      & \underline{0.640} & \underline{0.807} & 0.337 & 0.424 \\
MemSkill   & 0.579 & 0.779 & 0.179 & 0.247 \\

\textbf{HAGE}
           & \textbf{0.678} & \textbf{0.824}
           & \textbf{0.429} & \textbf{0.527}\\
\bottomrule
\end{tabular}
\end{table}

\subsection{System Efficiency Analysis}
\label{sec:rq3}

To evaluate the system efficiency of HAGE, we focus on two deployment-time metrics: (1) average token cost per query and (2) average query latency. We also report the average task score to compare the accuracy--efficiency trade-off across methods.

Table~\ref{tab:system_perf} reports the accuracy--efficiency comparison across different memory methods. HAGE achieves the highest average score among all methods. This improvement comes with a moderate increase in inference cost: HAGE uses $3.82$K tokens per query and reaches an average latency of $2.17$s. Compared with the most efficient high-performing baseline, HAGE trades a small amount of additional token and latency overhead for a clear improvement in average score.

Overall, the results suggest that HAGE provides a favorable accuracy--efficiency trade-off. It achieves the best task performance while keeping token consumption and latency within the same order of magnitude as other retrieval-based memory methods.

\begin{table}[t]
\centering
\small
\setlength{\tabcolsep}{5pt}
\caption{Accuracy--efficiency trade-off on the LoCoMo benchmark. Average score is evaluated by LLM-as-a-Judge, while token consumption and latency measure inference-time cost. Best and second-best results are highlighted in bold and underlined, respectively.}
\label{tab:system_perf}
\begin{tabular}{lccc}
\toprule
\textbf{Method} & \textbf{Avg. Score} & \textbf{Tokens/Query (K)} & \textbf{Latency (s)} \\
\midrule
A-MEM               & 0.580 & \underline{2.62} & 2.26 \\
MemoryOS            & 0.553 & 4.76 & 32.68 \\
Nemori              & 0.590 & 3.46 & 2.59 \\
MAGMA               & \underline{0.700} & 3.37 & \underline{1.72} \\
MemSkill            & 0.501 & \textbf{0.92} & \textbf{1.46}  \\

\textbf{HAGE}  & \textbf{0.739} & 3.82 & 2.17 \\
\bottomrule
\end{tabular}
\end{table}

\begin{table}[!t]
\centering
\small
\caption{Breakdown analysis on the performance impact of different schemes in HAGE.}
\label{tab:ablation}
\begin{tabular}{l|cc}
\toprule
HAGE schemes & {Judge} & {F1} \\
\midrule
Static Edge              & 0.698 & 0.462 \\
LLM Scorer Edges         & 0.712 & 0.500 \\
Trainable Edge           & 0.724 & 0.514 \\
Trainable Router         & 0.713 & 0.502 \\
\midrule
\textbf{HAGE}            & \textbf{0.739} & \textbf{0.548} \\
\bottomrule
\end{tabular}
\end{table}

\subsection{Effect of Learned Edges and Routing}

Table~\ref{tab:ablation} analyzes the contribution of different HAGE components. Static edges achieve a Judge score of 0.698, showing that the underlying graph structure is useful but insufficient when traversal relies on fixed edge semantics. LLM-scored edges improve the score to 0.712, and trainable edges further improve it to 0.724, indicating that query-aware and learned edge representations provide stronger retrieval signals. The trainable-router variant also improves over static edges, suggesting that adaptive traversal decisions are important for selecting useful evidence.

The full HAGE model performs best across all metrics, reaching 0.739 Judge, 0.548 F1. These gains suggest that learned edge representations and trainable routing are complementary: edge learning captures query-dependent relational usefulness, while router learning determines how to exploit these relational signals during traversal. This explains why jointly optimizing both components outperforms using static edges, LLM-scored edges, or a trainable router alone.

\section{Conclusion}
\label{sec:conclusion}
We present HAGE, a weighted multi-relational memory framework that formulates agentic memory retrieval as query-conditioned traversal over dynamic relational graphs. By coupling relation-aware graph traversal with reinforcement learning-based optimization of routing policies and edge representations, HAGE enables memory retrieval to adapt to both query intent and downstream task feedback. Empirical results show that HAGE improves long-horizon reasoning accuracy and offers a favorable accuracy-efficiency trade-off compared to state-of-the-art agentic memory systems. These findings suggest that dynamic, trainable, and relation-aware memory structures offer a promising foundation for more capable LLM agents.

\section*{Limitations}

HAGE has several limitations that scope the interpretation of our results.

\textbf{Benchmark coverage.} Our evaluation covers two benchmarks---LoCoMo (long-term conversational memory) and HotpotQA (non-conversational multi-hop QA). While these represent complementary retrieval settings, results may not fully generalize to other memory-intensive tasks such as procedural or document-grounded reasoning.

\textbf{Dependence on LLM components.} Both query analysis (relation intent extraction) and evaluation (LLM-as-a-Judge) rely on instruction-tuned LLMs. This introduces cost and model-specific variability; the accuracy of the relation intent classifier directly affects the quality of query-conditioned edge features used during traversal.

\section*{Ethical Considerations}

Persistent memory systems inherently raise privacy concerns: agents that accumulate detailed user interaction histories may retain sensitive personal information beyond its intended scope. In personalized agent deployments, this could enable misuse if memory stores are accessed without user consent or appropriate safeguards. Additionally, RL-optimized retrieval policies may learn to surface information in ways that reflect biases present in training data. We encourage practitioners deploying memory-augmented agents to implement appropriate data retention policies and user-control mechanisms. On the positive side, HAGE contributes to the development of more capable long-horizon AI agents by enabling structured, relation-aware memory retrieval, with potential benefits for applications such as personal assistants, knowledge-intensive dialogue systems, and automated research agents.

All datasets and models used in this work are publicly available and used in accordance with their respective licenses (LoCoMo under CC BY-NC 4.0; HotpotQA under CC BY-SA 4.0; all-MiniLM-L6-v2 under Apache 2.0; GPT-4o-mini via the OpenAI API; Qwen2.5-3B under the Qwen Research License). No new datasets are introduced.

\bibliography{custom}

\appendix

\section{Related Work}
\label{app:relatedwork}

Recent surveys have characterized agentic memory from complementary perspectives, including brain-inspired memory taxonomies, forms--functions--dynamics frameworks, efficiency-oriented agent design, graph-based memory lifecycles, and empirical analyses of evaluation and system limitations~\citep{jia2026ai,hu2025memory,yang2026toward,yang2026graph,jiang2026anatomy}.
We organize related work along four axes that situate HAGE within this broader literature: context-window extension, retrieval-augmented generation, structured and graph-based agent memory, and learning memory access policies.

\textbf{Context-Window Extension.}
A direct line of work extends the effective context length of Transformers through modified attention or positional encodings~\citep{beltagy2020longformer,press2021train}. More recent efforts augment decoders with auxiliary memory modules~\citep{kang2025lm2} or global-memory-enhanced retrieval pipelines~\citep{qian2025memorag} to handle inputs that exceed even extended context windows. While these approaches mitigate the context-length bottleneck, they do not address the continual, multi-session write-back nature of agentic memory, where the memory store itself must evolve in response to new interactions.

\textbf{Retrieval-Augmented Generation.}
RAG~\citep{lewis2020retrieval} conditions generation on passages retrieved from a static external corpus. Subsequent work has extended this paradigm to long-context LLMs~\citep{jiang2024longrag}, multi-partition retrieval~\citep{wang2024m}, and optimized retrieval serving~\citep{jiang2025rago}. Classical RAG formulations typically assume a relatively static knowledge base and retrieval over externally provided documents, even though later extensions introduce iterative or multi-hop retrieval. Agentic settings require memory that is continuously updated and accessed through multi-hop reasoning chains---motivating the shift to Memory-Augmented Generation (MAG) systems that support dynamic read--write--update loops.

\textbf{Structured and Graph-Based Agent Memory.}
Beyond flat vector stores, a growing body of work organizes agent memory into structured representations to support richer reasoning. MemGPT~\citep{packer2024memgptllmsoperatingsystems} introduces an OS-style memory hierarchy with explicit paging. MemoryBank~\citep{zhong2024memorybank} and Nemori~\citep{nan2025nemori} focus on episodic memory construction with selective write-back. A-MEM~\citep{xu2025mem} adopts a Zettelkasten-inspired linking strategy for note-like memory units. MemoryOS~\citep{Kang2025}, Zep~\citep{rasmussen2025zep}, and Hippocampus~\citep{li2026hippocampus} propose persistent or scalable memory modules for multi-session agents. These systems improve memory persistence and organization, but many still retrieve memories through vector similarity, recency, salience, or manually specified control rules.

Graph-based memory architectures explicitly encode relational structure. GraphRAG~\citep{edge2024graphrag} builds entity-centric community graphs for global question answering over large corpora. AriGraph~\citep{anokhin2024arigraph} constructs knowledge-graph world models with evolving episodic structure, enabling relational reasoning for LLM agents. GAM~\citep{wu2026gam} proposes a hierarchical graph memory organized around Event-Nodes and Episode-Nodes, demonstrating that multi-level graph organization improves long-horizon retrieval. EMG~\citep{wang2024crafting} combines editable graph-structured memory with retrieval-augmented generation for personalized agents. While these systems design expressive relational memory structures, their retrieval mechanisms remain largely static---relying on fixed edge weights, type-level scoring heuristics, or single-shot similarity search---rather than learning to route queries dynamically.

\textbf{Learning Memory Access Policies.}
A smaller but growing body of work frames memory access as a learnable decision process rather than a fixed retrieval procedure. AgeMem~\citep{yu2026agentic} proposes a unified long- and short-term memory management framework trained with reinforcement learning to optimize memory operations end-to-end, demonstrating that downstream reward signals can guide when and what to retrieve. Mariot et al.~\citep{mariot2026reconstruct} reconceptualize memory access as an iterative, multi-step reconstruction process rather than a static lookup, arguing that relevant memory must often be assembled across multiple retrieval steps before it can inform generation. These works share HAGE's core motivation---that retrieval should be optimized rather than hand-designed---but differ in scope: they focus on memory management policies or flat retrieval, whereas HAGE specifically targets query-conditioned traversal over multi-relational graph structures with jointly trained edge representations and routing policies.

Taken together, prior work demonstrates steady progress in structuring agent memory and improving retrieval coverage. HAGE addresses the intersection of these threads: structured multi-relational graph memory combined with RL-based, query-conditioned routing that adapts both traversal behavior and edge representations to downstream task feedback.

\section{Implementation Details}
\label{app:implementation}

Each sample graph stores node embeddings of size $N \times 384$ using all-MiniLM-L6-v2~\citep{reimers2019sentence}, edge indices in COO format, integer edge-type labels, and an $E \times 4$ edge feature matrix. Training uses 5-fold cross-validation at the sample level, 20\% held out for validation, and the remaining 10\% strictly reserved as an unseen test set per fold. Each fold trains for 200 epochs with Adam~\citep{kingma2014adam}, using $\eta_{router}=10^{-3}$ and $\eta_{edge}=10^{-4}$.

The remaining hyperparameters are: discount factor $\gamma=0.99$, baseline decay $\beta=0.99$, anchor regularization $\lambda_{anchor}=1.0$, hop budget $H_{max}=5$, hit reward $R_{hit}=10.0$, step penalty $\lambda_{step}=0.05$, and timeout penalty $\lambda_{timeout}=1.0$.

\section{Prompt Library}
\label{app:prompts}

HAGE employs a sophisticated prompt strategy with three distinct types, each optimized for specific cognitive tasks within the memory pipeline.

\subsection{Structured Event Extraction Prompt}
We use a structured event extraction prompt to convert raw conversational turns into graph-compatible memory units. The prompt asks the model to identify salient entities, topics, relationships, temporal cues, and concise factual summaries. Instead of relying on free-form generation, the extractor returns a lightweight structured output that can be directly consumed by the memory construction pipeline.

\begin{tcolorbox}[colback=gray!10, colframe=black, title=\textbf{Prompt Template: Event Extraction}, breakable]
\small
\raggedright
\textbf{Role:} Extract structured memory information from conversational text.

\textbf{Input:}
\begin{itemize}[nosep, leftmargin=*]
    \item Speaker: \{speaker\}
    \item Text: \{text\}
    \item Optional context: \{prev\_summary\}
\end{itemize}

\textbf{Output fields:}
\begin{itemize}[nosep, leftmargin=*]
    \item Entities or concepts mentioned in the text.
    \item Main topic or theme of the utterance.
    \item Relevant relationships among entities or events.
    \item Key factual information to preserve in memory.
    \item Temporal expressions, if any.
    \item A short summary with speaker attribution.
\end{itemize}
\end{tcolorbox}

\subsection{Query-Adaptive QA Prompt}
During answer generation, the retrieved memory context is provided to a QA prompt together with the user query. The prompt is adapted according to the query type predicted by our router, allowing the system to emphasize different reasoning behaviors when needed. This design keeps the generation stage grounded in retrieved memory while allowing lightweight query-specific control without exposing task-specific prompt details.

\begin{tcolorbox}[colback=gray!10, colframe=black, title=\textbf{Prompt Template: Query-Adaptive QA}, breakable]
\small
\raggedright
\textbf{Role:} Answer the user query using the retrieved memory context.

\textbf{Input:}
\begin{itemize}[nosep, leftmargin=*]
    \item Retrieved context: \{context\}
    \item User question: \{question\}
    \item Query guidance: \{router\_generated\_instruction\}
\end{itemize}

\textbf{General instructions:}
\begin{itemize}[nosep, leftmargin=*]
    \item Ground the answer in the provided context.
    \item Return a concise answer when the query asks for a specific fact.
    \item State that the information is unavailable when the context does not support an answer.
    \item Follow the router-generated guidance when additional reasoning control is required.
\end{itemize}

\textbf{Answer:}
\end{tcolorbox}

\subsection{Evaluation Prompt (LLM-as-a-Judge)}
To ensure rigorous evaluation beyond simple n-gram overlapping, we employ a semantic scoring mechanism. The Judge LLM evaluates the alignment between the generated response and the ground truth using the following schema.

\begin{tcolorbox}[colback=gray!10, colframe=black, title=\textbf{System Prompt: Semantic Grader}, breakable]
\small
\raggedright
You are an expert evaluator assessing the semantic fidelity of a memory retrieval system. Score the \texttt{Candidate Answer} against the \texttt{Gold Reference} on a continuous scale [0.0, 1.0].

\textbf{Scoring Rubric:}
\begin{itemize}[nosep, leftmargin=*]
    \item \textbf{1.0 (Exact Alignment):} Captures all key entities, temporal markers, and causal relationships. Semantically equivalent.
    \item \textbf{0.8 (Substantially Correct):} Main point is accurate but lacks minor nuances or secondary details.
    \item \textbf{0.6 (Partial Match):} Contains valid information but misses key constraints (e.g., wrong date but correct event).
    \item \textbf{0.4 (Tangential):} Touches on the topic but misses the core information requirement.
    \item \textbf{0.2 (Incoherent):} Factually incorrect with only minimal topical overlap.
    \item \textbf{0.0 (Contradiction/Hallucination):} Completely unrelated or contradicts the ground truth.
\end{itemize}

\textbf{Evaluation Constraints:}
\begin{enumerate}[nosep, leftmargin=*]
    \item \textbf{Temporal Flexibility:} Accept relative time references (e.g., ``next Tuesday'') if they resolve to the same period as the Gold Reference.
    \item \textbf{Semantic Equivalence:} Prioritize informational content over lexical matching.
    \item \textbf{Adversarial Handling:} If the Gold Reference states ``Unanswerable'', the Candidate MUST explicitly state lack of information. Any hallucinated fact results in 0.0.
\end{enumerate}

\textbf{Input:} Question: \{question\} | Gold: \{gold\} | Candidate: \{generated\}\\
\textbf{Output:} JSON \texttt{\{"score": float, "reasoning": "concise explanation"\}}
\end{tcolorbox}

\section{Baseline Configurations}
\label{app:baselines}

To ensure a fair and rigorous comparison, we standardized the experimental environment across all systems. Specifically, we adhered to the following protocols:

\begin{itemize}
\item \textbf{Full Context Baseline:} We implemented a ``Full Context'' baseline where the entire available conversation history is fed directly into the LLM's context window (up to the 128k token limit of \texttt{gpt-4o-mini}). This serves as a ``brute-force'' reference to evaluate the model's native long-context capabilities without external retrieval mechanisms.
\item \textbf{Retrieval-Based Baselines:} For all baseline systems (e.g., AMem, Nemori, MemoryOS), we applied their official default hyperparameters and storage settings to reflect their standard out-of-the-box performance.
\item \textbf{Unified Backbone Model:} To eliminate performance variance caused by different foundation models, all systems utilized OpenAI's \texttt{gpt-4o-mini} for both retrieval reasoning and response generation.
\item \textbf{Unified Evaluation:} All system outputs were evaluated using the identical \textit{LLM-as-a-Judge} framework (also powered by \texttt{gpt-4o-mini} with temperature=0.0), as detailed in Appendix~\ref{app:prompts}.
\end{itemize}

\paragraph{Dataset Statistics.}
We conducted a comprehensive evaluation on the full LoCoMo benchmark, testing across all five cognitive categories to assess varying levels of retrieval complexity. The detailed distribution of query types is presented in Table~\ref{tab:locomo_stats}.

\begin{table}[h]
\centering
\small
\caption{Distribution of query categories in the LoCoMo benchmark used for evaluation.}
\label{tab:locomo_stats}
\begin{tabular}{l r}
\toprule
\textbf{Query Category} & \textbf{Count} \\
\midrule
Single-Hop Retrieval & 841 \\
Adversarial & 446 \\
Temporal Reasoning & 321 \\
Multi-Hop Reasoning & 282 \\
Open Domain & 96 \\
\midrule
\textbf{Total Samples} & \textbf{1,986} \\
\bottomrule
\end{tabular}
\end{table}

\section{Dataset and Model Licenses}
\label{app:licenses}

We use the following publicly available datasets and models in our experiments:

\begin{itemize}[leftmargin=*, itemsep=0pt, topsep=2pt]
    \item \textbf{LoCoMo}~\citep{maharana2024evaluating}: Released under CC BY-NC 4.0.
    \item \textbf{HotpotQA}~\citep{yang2018hotpotqa}: Released under CC BY-SA 4.0.
    \item \textbf{all-MiniLM-L6-v2}~\citep{reimers2019sentence}: Released under Apache License 2.0.
    \item \textbf{GPT-4o-mini}~\citep{hurst2024gpt}: Accessed via the OpenAI API under OpenAI's Terms of Service.
    \item \textbf{Qwen2.5-3B}~\citep{yang2025qwen3}: Released under the Qwen Research License Agreement.
\end{itemize}

All datasets are used for research purposes consistent with their respective licenses. No new datasets are introduced in this work.

\end{document}